\documentclass{article} 
\pdfoutput=1
\usepackage{iclr2018_workshop,times}
\usepackage{url}
\usepackage{amsmath}
\usepackage{algorithm}
\usepackage{graphicx}
\usepackage{subfig}
\usepackage[noend]{algpseudocode}
\usepackage[acronym,smallcaps,nowarn,section,nonumberlist]{glossaries}
\glsdisablehyper

\makeatletter
\def\BState{\State\hskip-\ALG@thistlm}
\makeatother

\title{Decoding Decoders: \\
Finding Optimal Representation Spaces
\\ For Unsupervised Similarity Tasks}

\author{Vitalii Zhelezniak, Dan Busbridge, April Shen, Samuel L. Smith\thanks{Now at Google Brain.}\; \& Nils Y. Hammerla \\
babylon health, 60 Sloane Avenue, London, SW3 3DD, United Kingdom \\
\texttt{\{vitali.zhelezniak, dan.busbridge, april.shen,} \\
\texttt{\phantom{\{}nils.hammerla\}@babylonhealth.com, slsmith@google.com}
}

\usepackage{amssymb}                       
\newcommand{\mathbold}[1]{\ensuremath{\boldsymbol{\mathbf{#1}}}}

\newcommand{\mba}{\mathbold{a}}
\newcommand{\mbb}{\mathbold{b}}
\newcommand{\mbc}{\mathbold{c}}

\newcommand{\mbh}{\mathbold{h}}

\newcommand{\mbu}{\mathbold{u}}
\newcommand{\mbv}{\mathbold{v}}

\newcommand{\mbU}{\mathbold{U}}

\DeclareMathOperator*{\argmax}{arg\,max}
\DeclareMathOperator*{\argmin}{arg\,min}
\DeclareMathOperator*{\softmax}{softmax}

\newcommand{\mbbR}{\mathbb{R}}

\newcommand{\simdot}{\stackrel{\textrm{dot}}{\sim}}
\newcommand{\simrho}{\stackrel{\rho}{\sim}}
\usepackage{tikz}

\usetikzlibrary{fit,positioning,arrows.meta}
\tikzset{neuron/.style={shape=circle, minimum size=1.25cm, 
    inner sep=0, draw, font=\small}, io/.style={neuron, fill=gray!20}}

\makeglossaries

\newacronym{bow}{BOW}{bag-of-words}
\newacronym{cvae}{CVAE}{Conditional Variational Autoencoder}
\newacronym{dnn}{DNN}{Deep Neural Network}
\newacronym{glove}{GloVe}{Global Vectors for Word Representation}
\newacronym{gru}{GRU}{Gated Recurrent Unit}
\newacronym{gcn}{GCN}{Graph Convolutional Neural Networks}
\newacronym{kld}{KLD}{Kullback-Leibler Divergence}
\newacronym{idf}{IDF}{Inverse Document Frequency}
\newacronym{ig}{IG}{Information Gain}
\newacronym{isan}{ISAN}{Input Switched Affine Network}
\newacronym{lhs}{L.H.S.}{left hand side}
\newacronym{lstm}{LSTM}{Long Short Term Memory}

\newacronym{ml}{ML}{Machine Learning}
\newacronym{mle}{MLE}{Maximum Likelihood Estimator}
\newacronym{mnist}{MNIST}{Modified National Institute of Standards and Technology}

\newacronym{nlp}{NLP}{Natural Language Processing}
\newacronym{nn}{NN}{Neural Network}

\newacronym{oh}{OH}{One-Hot}

\newacronym{pdf}{PDF}{Probability Density Function}

\newacronym{rhs}{R.H.S.}{Right Hand Side}
\newacronym{rnn}{RNN}{Recurrent Neural Network}

\newacronym{sgd}{SGD}{Stochastic Gradient Descent}
\newacronym{sg}{SG}{SkipGram}
\newacronym{st}{ST}{SkipThought}
\newacronym{sts}{STS}{Semantic Textual Similarity}

\newacronym{termfreq}{TF}{Term Frequency}
\newacronym{tfidf}{TF-IDF}{Term Frequency-Inverse Document Frequency}

\newacronym{vae}{VAE}{Variational Autoencoder}

\begin{document}
\maketitle
\begin{abstract}
 Experimental evidence indicates that simple models outperform complex deep
 networks on many unsupervised similarity tasks.
  We provide a simple yet rigorous explanation for this behaviour
  by introducing the concept of an \emph{optimal representation space},
  in which semantically close symbols are mapped to representations that are close under
  a similarity measure induced by the model's objective function.
  In addition, we present a straightforward procedure that, without any retraining or
  architectural modifications, allows deep recurrent models to perform equally well
 (and sometimes better) when compared to shallow models.
  To validate our analysis, we conduct a set of consistent empirical evaluations and
  introduce several new sentence embedding models in the process.
 Even though this work is presented within the context of natural language
  processing, the insights are readily applicable to other domains that rely on distributed
  representations for transfer tasks.
\end{abstract}
\section{Introduction}

Distributed representations have played a pivotal role in the current success of
machine learning.
In contrast with the symbolic representations of classical AI, distributed
representation spaces can encode rich notions of semantic similarity in their
distance measures, allowing systems to generalise to novel inputs.
Methods to learn these representations have gained significant traction, in
particular for modelling words \citep{Mikolov2013}.
They have since been successfully applied to many other domains, including
images \citep{Le2011, Razavian2014a} and graphs \citep{Kipf2016a, Grover2016,
  Narayanan2017}.

Using unlabelled data to learn effective representations is at the forefront of modern machine learning research.
The \gls{nlp} community in particular has invested significant efforts in the construction \citep{Mikolov2013, Pennington2014, Bojanowski2016, Joulin2016a}, evaluation \citep{Baroni2014} and theoretical analysis \citep{NIPS2014_5477} of distributed representations for words.

Recently, attention has shifted towards the unsupervised learning of representations for larger pieces of text, such as
phrases \citep{Yin2015, Zhang2017a}, sentences \citep{Kalchbrenner2014, Kiros2015, Tai2015, Hill2016, Arora2017}, and entire paragraphs \citep{Le2014}.
Some of this work simply sums or averages constituent word vectors to obtain
a sentence representation \citep{Mitchell2010, Milajevs2014a, Wieting2016, Arora2017}, which is surprisingly effective.

Another line of research has relied on a \emph{sentence-level distributional hypothesis} \citep{Polajnar2015}, originally applied to words 
\citep{Harris1954}, which is an assumption that sentences which occur in similar contexts have a similar meaning.
Such models often use an encoder-decoder architecture \citep{Cho2014} to predict
the adjacent sentences of any given sentence.
Examples of such models include SkipThought \citep{Kiros2015}, which uses \glspl{rnn} for its encoder and decoders, and FastSent \citep{Hill2016}, which replaces the \gls{rnn}s with simpler \gls{bow} versions.

Models trained in an unsupervised manner on large text corpora are usually applied to \emph{supervised transfer tasks}, where the
representation for a sentence forms the input to a supervised classification problem, or to \emph{unsupervised similarity tasks}, where
the similarity (typically taken to be the cosine similarity) of two inputs is compared with corresponding human judgements of semantic similarity in order to inform some
downstream process, such as information retrieval.

Interestingly, some researchers have observed that deep complex models like SkipThought tend to do well on supervised transfer
tasks but relatively poorly on unsupervised similarity tasks, whereas for shallow log-linear models like FastSent the opposite is true
\citep{Hill2016, Conneau2017}.
It has been highlighted that this should be
addressed by analysing the geometry of the representation space \citep{Almahairi2015, Schnabel2015,
  Hill2016, Wieting2017}, however, to the best of our knowledge it has not been
systematically attempted.


In this work we attempt to address the observed performance gap on unsupervised
similarity tasks between representations produced by simple
models and those produced by deep complex models. Our main contributions are as follows:
\begin{itemize}
	\item We introduce the concept of a model's \emph{optimal representation space}, in
    which semantic similarity between symbols is mapped to a similarity measure
    between their corresponding representations, and that measure is induced by that model's objective function.
	\item We show that models with log-linear decoders are usually evaluated in their optimal space, while recurrent models are not.  This effectively explains the performance gap on unsupervised similarity tasks.
	\item We show that, when evaluated in their optimal space, recurrent models close that gap. We also provide a procedure for extracting
	this optimal space using the decoder hidden states.
	\item We validate our findings with a series of consistent empirical
    evaluations utilising a single publicly available codebase.
    \footnote{
      Code available at
      \href{https://www.github.com/Babylonpartners/decoding-decoders}{
        \texttt{github.com/Babylonpartners/decoding-decoders}}.}
\end{itemize}
\section{Optimal Representation Space}
\label{sec:optimal-representation-space}

We begin by considering a general problem of learning a conditional probability distribution $P_{\text{model}}(y\, |\, x)$ over
the output symbols $y\in \mathcal{Y}$ given the input symbols $x \in \mathcal{X}$.

\paragraph{Definition 1.} A space $\mathcal{H}$ combined with a similarity measure $\rho: \mathcal{H} \times \mathcal{H} \mapsto \mathbb{R}$
in which semantically close symbols $x_i, x_j \in \mathcal{X}$ have
representations $\mbh_{x_i}, \mbh_{x_j} \in \mathcal{H}$ that are close in $\rho$ is called a \emph{distributed representation space} \citep{Goodfellow2016a}.



In general, a distributed representation of a symbol $x$ is obtained via some function $\mbh_x = f(x; \theta_f)$, parametrised by weights $\theta_f$.
Distributed representations of the input symbols are typically found as the
layer activations of a \gls{dnn}.
One can imagine running all possible $x\in \mathcal X$ through a \gls{dnn} and
using the activations $\mbh_x$ of the $n^{th}$ layer as vectors in $\mathcal{H}_x$:
\begin{equation*}
\mathcal{H}_x = \left\{\mbh_x = \text{Activation}^{(n)}(x)\; | \; x \in \mathcal{X} \right\}.
\end{equation*}
The distributed representation space of the output symbols $\mathcal{H}_y$ can be obtained via some function $\mbh_y = g(y; \theta_g)$ that does not depend
on the input symbol $x$, e.g. a row of the softmax projection matrix that corresponds to the output $y$.

In practice, although $\mathcal{H}$ obtained in such a manner with a reasonable vector similarity $\rho$ (such as cosine or Euclidean
distance) forms a distributed representation space, there is no \emph{a priori} reason why an arbitrary choice of a similarity function would be
appropriate given $\mathcal{H}$ and the model's objective.
There is no analytic guarantee, for arbitrarily chosen $\mathcal{H}$ and $\rho$, that small changes in
semantic similarity of symbols correspond to small changes in similarity $\rho$ between their vector representations in $\mathcal{H}$ and vice versa,
unless such a requirement is induced by optimising the objective function.
This motivates Definition 2.

\paragraph{Definition 2.}
A space $\mathcal{H}$ equipped with a similarity measure $\rho$ such that
$\log P_{\text{model}}(y\, |\, x) \propto \rho \left(\mbh_y, \mbh_x \right)$ is called an \emph{optimal representation space}.

In words, if a model has an optimal representation space, the conditional log-probability of an output symbol $y$ given an input symbol $x$ is proportional
to the similarity $\rho(\mbh_y, \mbh_x)$  between their corresponding vector representations $\mbh_y, \mbh_x \in \mathcal{H}$.

For example, consider the following standard classification model
\begin{equation}
  \label{eq:softmax_model}
  P_{\text{model}}(y\, |\, x) = \frac{\exp{(\mbu_y\cdot\text{DNN}(x))}}{\sum_{y'}\exp{(\mbu_{y'}\cdot\text{DNN}(x))}}
\end{equation}
where $\mbu_y$ is the $y^{th}$ row of the output projection matrix $\mbU$.

If $\mathcal{H}_x = \left\{\text{DNN}(x)\; | \; x \in \mathcal{X} \right\}$ and $\mathcal{H}_y = \left\{\mbu_y\; | \; y \in \mathcal{Y} \right\}$, then $\mathcal{H} = \mathcal{H}_x \cup \mathcal{H}_y$
equipped with the standard dot product $\rho (\mbh_{1}, \mbh_{2}) = \mbh_{1} \cdot \mbh_{2}$ is an optimal representation space. Note that if the exponents of
\autoref{eq:softmax_model} contained Euclidean
distance, then we would find $\log P_{\text{model}}(y \,| \,x) \propto ||\mbu_y -
\text{DNN}(x)||_2$. The optimal representation space would then be equipped with Euclidean
distance as its optimal distance measure $\rho$. This easily extends to any other distance measures
desired to be induced on the optimal representation space.

Let us elaborate on why Definition 2 is a reasonable definition of an optimal space.
Let $x_1, x_2 \in \mathcal{X}$ be the input symbols and $y_1, y_2 \in \mathcal{Y}$ their
corresponding outputs.
Using
\begin{equation*}
 \mba \simrho \mbb
\end{equation*}
to denote that  $\mba$ and $\mbb$ are close under $\rho$, a reasonable model trained on a subset of $(\mathcal{X}, \mathcal{Y})$ will
ensure that $\mbh_{x_1} \simrho \mbh_{y_1}$ and $\mbh_{x_2} \simrho \mbh_{y_2}$.
If $x_1$ and $x_2$ are semantically close and
assuming semantically close input symbols have similar outputs, we also have that $\mbh_{x_1} \simrho \mbh_{y_2}$ and $\mbh_{x_2} \simrho \mbh_{y_1}$.
Therefore it follows that $\mbh_{x_1} \simrho \mbh_{x_2}$ (and $\mbh_{y_1} \simrho \mbh_{y_2}$).
Putting it differently, semantic similarity of input and output symbols translates into closeness of their distributed representations under $\rho$, in a way that is consistent with the model.


Note that any model $P_{\text{model}}(y\, |\, x)$
parametrised by a continuous function can be approximated by a function in the
form of \autoref{eq:softmax_model}. It follows that any model that produces a probability
distribution has an optimal representation space. Also note that the optimal space for the inputs does not necessarily have to come from the final layer
before the softmax projection but instead can be constructed from any layer, as
we now demonstrate.

Let $n$ be the index of the final activation before the softmax projection and let $k \in \{1,\ldots,n\}$.
We split the network into three parts:
\begin{equation}
\softmax \left(\mbU F_n\left(G_k(x)\right) \right)
\end{equation}
where $G_k$ contains first $k$ layers, $F_n$ contains the remaining $n-k$ layers and $\mbU$ is the softmax projection matrix.
Let the space for inputs $\mathcal{H}_x$ be defined as
\begin{equation*}
  \mathcal{H}_x
  = \left\{G_k(x)\; | \; x \in \mathcal{X} \right\}
\end{equation*}
and the space for outputs $\mathcal{H}_y$ defined as
\begin{equation*}
  \mathcal{H}_y
  = \left\{\mbu_y\; | \; y \in \mathcal{Y} \right\}.
\end{equation*}
Their union  $\mathcal{H} = \mathcal{H}_x \cup \mathcal{H}_y$ equipped with $\rho (\mbh_{1}, \mbh_{2}) = J (\mbh_1)\, \cdot J \,(\mbh_2)$
where
\begin{equation*}
J(\mathbf{h}) =
  \begin{cases}
    F_n(\mathbf{h}) & \quad \text{if } \mathbf{h} \in \mathcal{H}_x \\
    \mathbf{h}  & \quad \text{otherwise} \\
  \end{cases}
\end{equation*}
is again an optimal representation space.

%

\section{Optimal Sentence Representation Space}
\label{sec:similarity}

For the remainder of this paper, we focus on unsupervised models for learning distributed representations of sentences,
an area of particular interest in \gls{nlp}.

\subsection{Background}
\label{subsec:similarity-background}

Let $S = (s_1, s_2, \ldots, s_N)$ be a corpus of contiguous sentences where each sentence
$s_i = w_{s_i}^1 w_{s_i}^2 \ldots w_{s_i}^{\tau_{s_i}}$ consists of words from a pre-defined vocabulary $V$ of size $|V|$.

We transform the corpus into a set of pairs $D = \{(s_i, c_i)\}_{i=1}^N$, where $s_i \in S$ and $c_i$ is a context of $s_i$.
The context usually (but not necessarily) contains some number of surrounding sentences of $s_{i}$, e.g. $c_i = (s_{i-1}, s_{i+1})$.

We are interested in modelling the probability of a context $c$ given a sentence $s$.
In general
\begin{equation}
  \label{eq:autoregressive_factorisation}
 P_{\text{model}}(c\, |\, s;\theta)
  = \prod_{t=1}^{\tau_{c}} P_{\text{model}}(w_c^t\; |\; w_c^{t-1},\ldots,w_c^1,s;\theta).
\end{equation}
One popular way to model $P(c\, |\, s)$ for sentence-level data is suggested by the encoder-decoder framework.
The encoder $\mathcal{E}$ produces a fixed-length vector representation $\mbh^{\mathcal{E}}_s = \mathcal{E}(s)$ for a sentence $s$
and the decoder gives a context prediction $\hat{c} = \mathcal{D}(\mbh^{\mathcal{E}}_s)$ from that representation.

Due to a clear architectural separation between $\mathcal{E}$ and $\mathcal{D}$, it is common to take $\mbh^{\mathcal{E}}_s$ as a representation of a sentence $s$ in
the downstream tasks.
Furthermore, since $\mbh^{\mathcal{E}}_s$ is usually encoded as a vector, such representations are often compared via simple similarity measures,
such as dot product or cosine similarity.

\subsection{Log-Linear Decoders}
\label{subsec:similarity-log-linear-decoder}

We first consider encoder-decoder architectures with a log-linear \gls{bow} decoder for the context.
Let $\mbh_i = \mathcal{E}(s_i)$ be a sentence representation of $s_i$ produced by some encoder $\mathcal{E}$.
The nature of $\mathcal{E}$ is not important for our analysis; for concreteness,
the reader can consider a model such as FastSent \citep{Hill2016}, where $\mathcal{E}$ is a \gls{bow} (sum) encoder.



In the case of the log-linear \gls{bow} decoder, words are conditionally independent of the previously occurring sequence, thus \autoref{eq:autoregressive_factorisation} becomes
\begin{equation}
  P_{\text{model}}(c_i|s_i;\theta)
  =\prod_{w\in c_i}P_{\text{model}}(w|s_i;\theta)
  = \prod_{w\in c_i} \frac{\exp{(\mbu_w \cdot \mbh_i)}}{\sum_{w'\in V}\exp{(\mbu_{w'} \cdot \mbh_i)}}.
\end{equation}
where $\mbu_w\in\mbbR^{d}$ is the output word embedding for a word $w$ and $\mbh_i$ is the encoder output. (Biases are omitted for brevity.)

The objective is to maximise the model probability of contexts $c_i$ given sentences $s_i$ across the corpus $D$,
which corresponds to finding the \gls{mle} for the trainable parameters $\theta$:
\begin{equation}
  \begin{split}
    \label{eq:mle_objective_log_linear}
    \theta_{\text{MLE}}
    & = \argmax_\theta \prod_{(s_i,c_i)\in D} P_{\text{model}}(c_i|s_i;\theta).
\end{split}
\end{equation}

By switching to the negative log-likelihood and inserting the above expression, we arrive at the following optimisation problem:
\begin{equation}
  \label{eq:mle_log_linear}
  \theta_{\text{MLE}}
  = \argmin_\theta \left[- \sum_{(s_i,c_i)\in D} \left(\sum_{w\in c_i}\mbu_w\cdot\mbh_i
  + |c_i|\log{\sum_{w'\in V}\exp{(\mbu_{w'}\cdot\mbh_i)}}\right) \right].
\end{equation}
Noticing that
\begin{equation}
  \label{eq:dot_product_additivity}
    \sum_{w\in c_i}\mbu_w\cdot\mbh_i = \left( \sum_{w\in c_i} \mbu_w \right)\cdot \mbh_i = \mbc_i \cdot \mbh_i,
\end{equation}
we see that the objective in \autoref{eq:mle_log_linear} forces the sentence representation $\mbh_i$ to be similar under dot product to its context representation $\mbc_i$, which is simply the sum of the output embeddings of the context words.
Simultaneously, output embeddings of words that do not appear in the context of a sentence are forced to be dissimilar to its representation.

Using $\simdot$ to denote \emph{close under dot product}, we find that if two sentences $s_i$ and $s_j$ have similar contexts, then $\mbh_i \simdot \mbc_j$ and $\mbh_j \simdot \mbc_i$.
The objective function in \autoref{eq:mle_log_linear} ensures that $\mbh_i \simdot \mbc_i$ and $\mbh_j \simdot \mbc_j$.
Therefore, it follows that $\mbh_i \simdot \mbh_j$.

Putting it differently, sentences that occur in related contexts are assigned
representations that are similar under the dot product.
Hence we see that the encoder output equipped with the dot product constitutes an optimal representation space as defined in \autoref{sec:optimal-representation-space}.



\subsection{Recurrent Sequence Decoders}
\label{subsec:similarity-sequence_decoder}

Another common choice for the context decoder is an \gls{rnn} decoder
\begin{align}
  \mbh^t & = \text{RNNCell}\left(\mbv^t, \mbh^{t-1} \right), \quad
  \mbh^0  = \mbh_i
\end{align}
where $\mbh_i = \mathcal{E}(s_i)$ is the encoder output. The specific structure of $\mathcal{E}$ is again not important for our analysis.
(When $\mathcal{E}$ is also an \gls{rnn}, this is similar to SkipThought \citep{Kiros2015}.)

The time unrolled states of decoder are converted to probability distributions over the vocabulary,
conditional on the sentence $s_i$ and all the previously occurring words.
\autoref{eq:autoregressive_factorisation} becomes
%

\begin{equation}
  P_{\text{model}}(c_{i}|s_i;\theta)
  = \prod_{t=1}^{\tau_{c_{i}}} P_{\text{model}}(w^t|w^{t-1},\ldots,w^1,s_i;\theta)
  = \prod_{t=1}^{\tau_{c_{i}}} \frac{\exp{(\mbu_{w^t} \cdot \mbh^t)}}{\sum_{w'\in V}\exp{(\mbu_{w'} \cdot \mbh^t)}}
\end{equation}

\begin{figure}[t]
  \centering
    \includegraphics[width=1\textwidth]{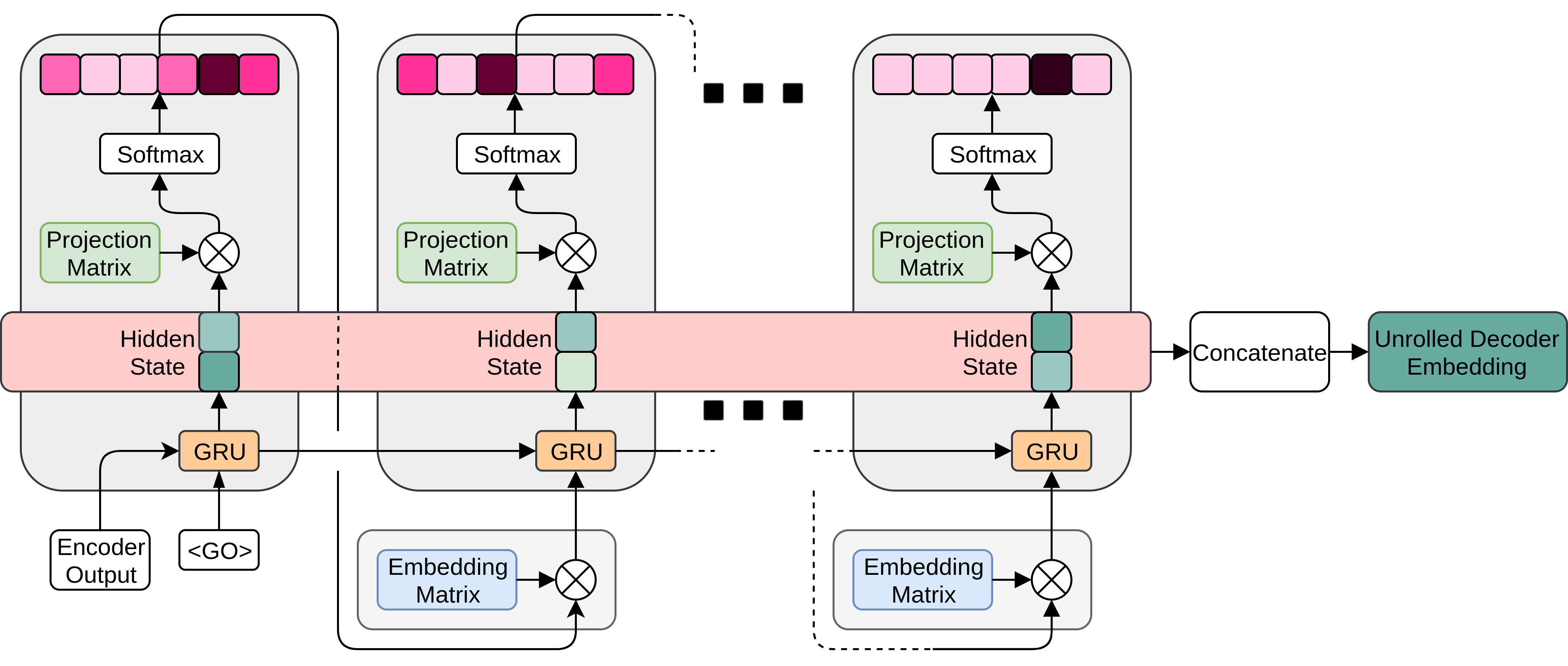}
    \caption{\small Unrolling an \gls{rnn} decoder at inference time. The initial hidden state for the decoder is typically the encoder output,
    either the recurrent cell final
      state for an \gls{rnn} encoder, or the sum of the input word embeddings for a BOW
      encoder.
      At the first time step, a learned \texttt{<GO>} token is presented as the input.
      In subsequent time steps, a probability-weighted sum over word vectors is used. The decoder is then unrolled for a
      fixed number of steps. The hidden states are then
      concatenated to produce the unrolled decoder embedding. In the models evaluated in
      \autoref{subsec:experiments-setup}, this process is
      performed for the \gls{rnn} corresponding to the previous and next sentences.
      The sentence representation is then taken as the
      concatenation across both \glspl{rnn}.}
    \label{fig:unrolling}
    \vspace{-1em}
\end{figure}

Similarly to \autoref{eq:mle_log_linear}, \gls{mle} for the model parameters $\theta$ can be found as
\begin{equation}
  \label{eq:mle_rnn}
  \theta_{\text{MLE}}
  = \argmin_\theta
  \left[ -
    \sum_{(s_i,c_i)\in D}
    \sum_{t=1}^{\tau_{c_{i}}} \left(
        \mbu_{w^t}\cdot\mbh^t
        + \log{\sum_{w'\in V}\exp{(\mbu_{w'}\cdot\mbh^t)}}
    \right)
  \right].
\end{equation}
Using $\oplus$ to denote \emph{vector concatenation}, we note that
\begin{equation}
  \label{eq:sequence_concat}
    \sum_{t=1}^{\tau_{c_{i}}}
        \mbu_{w^t}\cdot\mbh^t
    = \left( \bigoplus_{t=1}^{\tau_{c_{i}}} \mbu_{w^t} \right)
    \cdot \left( \bigoplus_{t=1}^{\tau_{c_{i}}} \mbh^t \right)
    = \mbc_i \cdot \mbh^{\mathcal{D}}_i,
\end{equation}
where the sentence representation $\mbh^{\mathcal{D}}_i$ is now an \emph{ordered} concatenation
of the hidden states of the decoder and the context representation $\mbc_i$ is
an \emph{ordered} concatenation of the output embeddings of the context words.
Hence we come to the same conclusion as in the log-linear case, except we have order-sensitive representations as opposed to unordered ones.  As before, $\mbh^{\mathcal{D}}_i$ is forced to be similar
to the context $\mbc_i$ under dot product, and is made dissimilar to sequences of $\mbu_{w'}$ that do not appear in the context.

The ``transitivity'' argument from \autoref{subsec:similarity-log-linear-decoder} remains intact, except
the length of decoder hidden state sequences might differ from sentence to sentence.
To avoid this problem, we can formally treat them as infinite-dimensional vectors in $\ell^2$ with
only a finite number of initial components occupied by the sequence and the rest set to zero.
Alternatively, we can agree on the maximum sequence length, which in practice can be determined from the training corpus.

Regardless, the above space of unrolled concatenated decoder states, equipped with dot product, is the optimal representation space for models
with recurrent decoders. Consequently, this space could be a much better candidate for unsupervised similarity tasks.

We refer to the method of accessing the decoder states at every time step as
\emph{unrolling the decoder}, illustrated in \autoref{fig:unrolling}. Note that accessing the decoder output does not require re-architecting or retraining the model,
yet gives a potential performance boost on unsupervised similarity tasks almost for free. We will demonstrate the effectiveness of this technique empirically in \autoref{subsec:experiments-results}.




\vspace{-0.5em}
\section{Experimental setup}
\label{subsec:experiments-setup}
\vspace{-0.5em}

\begin{figure}[t]
  \centering
    
    {\includegraphics[width=\textwidth]{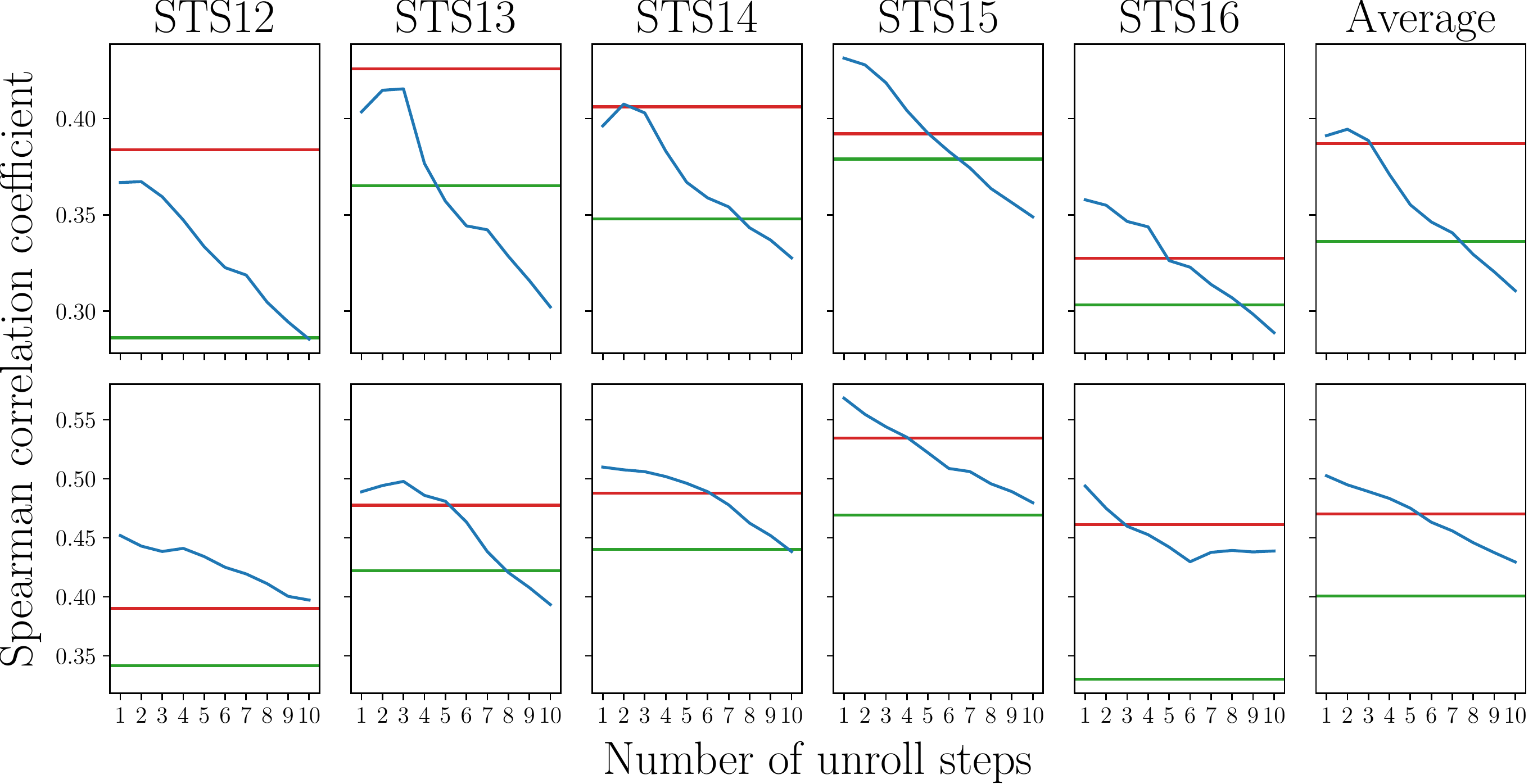}}
    \caption{\small Performance on the STS tasks depending on the number of unrolled hidden states of the decoders,
    using dot product as the similarity measure.
    The top row presents results for the \gls{rnn} encoder and the bottom row for the \gls{bow} encoder.
    \textbf{Red:} Raw encoder output with \gls{bow} decoder. \textbf{Green:} Raw encoder output with \gls{rnn} decoder.
    \textbf{Blue:} Unrolled \gls{rnn} decoder output.
    Independent of the encoder architecture, unrolling even a single state of
    the decoder always outperforms the raw encoder output with \gls{rnn} decoder, and
    almost always outperforms the raw encoder output with \gls{bow} decoder for some
    number of unrolls.
    }
    \label{fig:decoder-unroll-dot}
    \vspace{-1em}
\end{figure}

We have seen in \autoref{sec:optimal-representation-space} that the optimal representation space for a given model depends on the choice of decoder architecture.
To support this theory, we train several encoder-decoder architectures for sentences with the decoder types analysed in \autoref{sec:similarity},
and evaluate them on downstream tasks using both their optimal space and the standard space of the encoder output
as the sentence representations.

\vspace{-1em}
\paragraph{Models and training.} Each model has an encoder for the current sentence, and decoders for the previous and next sentences.
As our analysis is independent of encoder type, we train and evaluate models with \gls{bow} and \gls{rnn} encoders, 
two common choices in the literature for sentence representation learners \citep{Hill2016, Kiros2015}.
The \gls{bow} encoder is the sum of word vectors \citep{Hill2016}. The \gls{rnn}
encoder and decoders are \glspl{gru} \citep{Cho2014}.

Using the notation ENC-DEC, we train RNN-RNN, RNN-BOW, BOW-BOW, and BOW-RNN models.
For each encoder-decoder combination, we test several methods of extracting sentence representations to be used in the downstream tasks.
First, we use the standard choice of the final output of the encoder as the sentence representation.
In addition, for models that have RNN decoders, we unroll between 1 and 10 decoder hidden states.
Specifically, when we unroll $n$ decoder hidden states, we take the first $n$ hidden
states from each of the decoders and concatenate them in order to get the resulting sentence representation.
We refer to these representations as *-RNN-concat.

All models are trained on the Toronto Books Corpus \citep{Zhu2015}, a dataset of 70 million ordered sentences from over 7,000 books.
The sentences are pre-processed such that tokens are lower case and splittable by space.

\vspace{-1em}
\paragraph{Evaluation tasks.}  We use the SentEval tool \citep{Conneau2017} to benchmark sentence embeddings on
both supervised and unsupervised transfer tasks.
The supervised tasks in SentEval include paraphrase identification (MSRP) \citep{Dolan2004}, movie review sentiment (MR) \citep{Pang2005}, product review sentiment (CR), \citep{Hu2004}), subjectivity (SUBJ) \citep{Pang2004}, opinion polarity (MPQA) \citep{Wiebe2005}, and question type (TREC) \citep{Voorhees2002, Li2002}.  In addition, there are two supervised tasks on the SICK dataset, entailment and relatedness (denoted SICK-E and SICK-R) \citep{Marelli2014}.  For the supervised tasks, SentEval trains a logistic regression model with 10-fold cross-validation using the model's embeddings as features.

The unsupervised \gls{sts} tasks are STS12-16 \citep{Cer2017, Agirre2012, Agirre2013a, Agirre2014, Agirre2015, Agirre2016}, which are scored without training a new supervised model; in other words, the embeddings are used to directly compute similarity,
which is compared to human judgements of semantic similarity.
We use dot product to compute similarity as indicated by our analysis; results and discussion using cosine similarity, which is canonical in the literature,
are presented in \autoref{sec:appendix-cos}.
For more details on all tasks and the evaluation strategy, see \cite{Conneau2017}.

\vspace{-1em}
\paragraph{Implementation and hyperparameters.}
Our goal is to study how different decoder types affect the performance of sentence embeddings on various tasks.
To this end, we use identical hyperparameters and architecture for each model (except encoder and decoder types),
allowing for a fair head-to-head comparison.
Specifically, for RNN encoders and decoders we use a single layer GRU with layer normalisation \citep{Ba2016}.
All the weights (including word embeddings) are initialised uniformly over $[-0.1, 0.1]$ and trained with Adam \citep{Kingma2014a} without weight decay or dropout.
Sentence length is clipped or zero-padded to 30 tokens and end-of-sentence tokens are used throughout training and evaluation.
Following \cite{Kiros2015}, we use a vocabulary size of $20k$ with vocabulary expansion,  $620$-dimensional word embeddings,
and $2400$ hidden units in all \glspl{rnn}.

\vspace{-0.5em}
\section{Results}
\label{subsec:experiments-results}
\vspace{-0.5em}

\begin{table}[t]
\centering
\small  
\caption{\small Performance of different architectures and sentence representations on
  unsupervised similarity tasks using dot product as the similarity measure.
  On each task, the highest performing setup for each
  encoder type is highlighted in bold and the highest performing setup overall
  is underlined.
  All reported values indicate Pearson/Spearman correlation
  coefficients for the task.
  In the case of both encoder types, unrolling the \gls{rnn} decoder using
  the concatenation of the decoder hidden states (*-\gls{rnn}-concat) dramatically improves the performance
  across all tasks compared to using the raw encoder output (*-\gls{rnn}), validating the
  theoretical justification presented in
  \autoref{subsec:similarity-sequence_decoder}.
}
\label{tab:similarity-dot}
\resizebox{\textwidth}{!}{
\begin{tabular}{cc|ccccc}
\hline
  \textbf{Encoder} & \textbf{Decoder}  &                     \textbf{STS12}                      &                     \textbf{STS13}                      &                     \textbf{STS14}                      &                     \textbf{STS15}                      &           \textbf{STS16}           \\
\hline \hline
     & BOW      &                 $0.286/\mathbf{0.384}$                  &                      $0.381/\mathbf{0.426}$                      &                      $0.365/\mathbf{0.406}$                      &                      $0.262/0.392$                      &           $0.260/0.328$            \\
     RNN & RNN      &                      $0.267/0.286$                      &                      $0.371/0.365$                      &                      $0.357/0.348$                      &                      $0.379/0.379$                      &           $0.313/0.303$            \\
  & RNN-concat  &                 $\mathbf{0.335}/0.359$                  &                      $\mathbf{0.411}/0.415$                      &                      $\mathbf{0.413}/0.403$                      &                      $\mathbf{0.414}/\mathbf{0.419}$                      &  $\mathbf{0.326}/\mathbf{0.347}$   \\ \hline
     & BOW      &                      $0.351/0.390$                      &                      $0.418/0.478$                      &                      $0.442/0.488$                      &                      $0.455/0.535$                      & $0.370/\underline{\mathbf{0.461}}$ \\
     BOW & RNN      &                      $0.310/0.342$                      &                      $0.365/0.422$                      &                      $0.396/0.440$                      &                      $0.412/0.469$                      &           $0.281/0.330$            \\
  & RNN-concat  & $\underline{\mathbf{0.422}}/\underline{\mathbf{0.438}}$ & $\underline{\mathbf{0.478}}/\underline{\mathbf{0.498}}$ & $\underline{\mathbf{0.498}}/\underline{\mathbf{0.506}}$ & $\underline{\mathbf{0.512}}/\underline{\mathbf{0.544}}$ & $\underline{\mathbf{0.402}}/0.460$ \\
\hline
\end{tabular}}
\vspace{-0.75em}
\end{table}

Performance of the unrolled models on the \gls{sts} tasks is presented in
\autoref{fig:decoder-unroll-dot}.
We note that unrolling even a single state of the \gls{rnn} decoder always improves the
performance over the raw encoder output.  In addition, the unrolled \gls{rnn} representation
is often able to outperform raw encoder output with \gls{bow} decoder for some number of
hidden states, providing further evidence of the efficacy of this method.

We observe that the performance tends to peak around 2-3 hidden states and fall off afterwards.
In principle, one might expect the peak to be around the average sentence length of the corpus.
A possible explanation of this behaviour is the ``softmax drifting effect''.
As there is no context available at inference time, we generate the word embedding for the next time step
using the softmax output from the previous time step (see
\autoref{fig:unrolling}). Given that for any sentence, there is no single correct
context, the probability distribution over the next words in that context will
be potentially multi-modal. This will produce inputs for the decoder that
diverge from the inputs it expects (i.e. word vectors for the vocabulary).  Further work is needed to understand this and other
possible causes in detail.

\begin{table}[t]
\centering
\caption{\small Performance of different architectures and sentence representations on
  supervised transfer tasks. On each task, the highest performing setup for each
  encoder type is highlighted in bold and the highest performing setup overall
  is underlined. All reported values indicate test accuracy on the task, except
  for SICK-R where we report the Pearson correlation with human-provided
  scores. Note that the analysis in \autoref{sec:similarity} is not readily applicable here,
as instead of using a similarity measure in the representation space directly,
the supervised transfer tasks train an entirely new model on top of the chosen
representation. However, the results do indicate that unrolling \gls{rnn} decoders
could be a reasonable choice even for supervised tasks.}
\label{tab:transfer}
\small
\resizebox{\textwidth}{!}{
\begin{tabular}{cc|ccccccccc}
\hline
  \textbf{Encoder} & \textbf{Decoder}   &         \textbf{MR}          &         \textbf{CR}          &        \textbf{MPQA}         &        \textbf{SUBJ}         &         \textbf{SST}         &        \textbf{TREC}         &        \textbf{MRPC}         &       \textbf{SICK-R}       &       \textbf{SICK-E}        \\
\hline\hline
    & BOW      &           $75.78$            &           $79.34$            &           $86.25$            &           $90.77$            &           $81.99$            &           $84.60$            &           $70.55$            &           $0.80$            &           $78.81$            \\
   RNN & RNN    & $\mathbf{77.06}$ &       $81.77$       & $\mathbf{88.59}$ &       $\mathbf{92.56}$       & $\underline{\mathbf{82.65}}$ &       $86.60$       &           $\mathbf{71.94}$            &           $\mathbf{0.83}$            & $\mathbf{81.10}$ \\
  & RNN-concat &           $76.20$            &           $\textbf{82.07}$            &           $85.96$            &           $91.80$            &           $80.83$            &           $\textbf{87.20}$            & $71.59$ &       $0.82$       &           $80.35$            \\ \hline
 &    BOW      &       $76.16$       &           $81.14$            &       $87.03$       & $92.77$ &       $81.66$       &           $84.20$            &           $71.07$            & $0.84$ &       $80.58$       \\
  BOW &  RNN    &           $76.05$            & $\underline{\mathbf{82.07}}$ &           $85.80$            &           $92.13$            &           $80.83$            & $87.20$ &       $72.99$       &           $0.82$            &           $78.87$            \\
  & RNN-concat &           $\underline{\textbf{77.27}}$            &           $82.04$            &           $\underline{\textbf{88.74}}$            &           $\underline{\textbf{92.88}}$            &           $\textbf{81.82}$            &           $\underline{\textbf{89.60}}$            & $\underline{\mathbf{73.68}}$ &       $\underline{\mathbf{0.85}}$       &           $\underline{\mathbf{82.26}}$            \\ \hline
\end{tabular}}
\vspace{-0.75em}
\end{table}

Performance across unsupervised similarity tasks is presented in \autoref{tab:similarity-dot}
and performance across supervised transfer tasks is presented in
\autoref{tab:transfer}. 
 For the unrolled architectures, in these tables we report on the one that performs best on the \gls{sts} tasks.
 In addition, see \autoref{sec:appendix-decoder} for a comparison with the original
SkipThought results from the literature, and \autoref{sec:appendix-cos} for results
using cosine similarity rather than dot product as the similarity measure in \gls{sts} tasks.

In the case of the unsupervised similarity tasks (\autoref{tab:similarity-dot}), note that architectures
evaluated in their optimal space, including unrolled \gls{rnn} and \gls{bow} decoders,
consistently outperform those evaluated in a sub-optimal space,
as our analysis in \autoref{sec:similarity} predicts.

When we look at the performance on supervised transfer tasks in \autoref{tab:transfer}, combined with the similarity
results in \autoref{tab:similarity-dot}, we see that the notion that models cannot be good at both supervised
transfer and unsupervised similarity tasks
needs refining; for example,
\gls{rnn}-\gls{rnn} achieves strong performance on supervised transfer,
while \gls{rnn}-\gls{rnn}-concat achieves strong performance on
unsupervised similarity.
In general, our results indicate that a single model may be able to perform well on different
downstream tasks, provided that the representation spaces chosen for each task
are allowed to differ.

Curiously, the unusual combination of a \gls{bow} encoder and concatenation of the \gls{rnn} decoders
leads to the best performance on most benchmarks, even slightly exceeding that
of some supervised models on some tasks \citep{Conneau2017}. This architecture may be worth
investigating.

\vspace{-0.5em}
\section{Conclusion}
\label{sec:conclusion}
\vspace{-0.5em}
In this work, we introduced the concept of an optimal representation space, where semantic
similarity directly corresponds to distance in that space, in order to shed light on the performance
gap between simple and complex architectures on downstream tasks.
In particular, we studied encoder-decoder architectures and how the representation space induced
by the encoder outputs relates to the training objective of the model.

Although the space of encoder outputs, equipped with dot product as the similarity function, has been typically taken
as the representation space regardless of what type of decoder is used, it turns out that this is only optimal
in the case of \gls{bow} decoders but not \gls{rnn} decoders. This yields a simple explanation for the observed
performance gap between these architectures, namely that the former has been evaluated in its optimal
representation space, whereas the latter has not.

Furthermore, we showed that any neural network that outputs a probability distribution has an optimal representation space.
Since an \gls{rnn} does produce a probability distribution, we analysed its objective function which motivated
a procedure of unrolling the decoder. This simple method allowed us to extract
representations that are provably optimal under dot product, without needing to retrain the model.



We then validated our claims by comparing the empirical performance of different architectures across transfer tasks.
In general, we observed that unrolling even a single state of
the \gls{rnn} decoder always outperforms the raw encoder output with \gls{rnn} decoder, and
almost always outperforms the raw encoder output with \gls{bow} decoder for some
number of unrolls.
This indicates different vector embeddings
can be used for different downstream tasks depending on what type of representation space is most
suitable, potentially yielding high performance on a variety of tasks from a single trained model.

Although our analysis of encoder-decoder architectures only considered \gls{bow} and
\gls{rnn} components, others such as convolutional \citep{Gehring2017} and graph-based \citep{Kipf2016a} ones
are more appropriate for some tasks. Additionally, although we focus on Euclidean
spaces, it has been shown that hyperbolic spaces \citep{Nickel2017}, complex-valued vector spaces \citep{Trouillon2016} and
spinor spaces \citep{Kanjamapornkul2017} all have beneficial modelling properties.
In each case, it is not necessarily true that evaluations are currently leveraging an optimal representation space
for each model.  However, as we showed with the \gls{rnn} decoder, analysing the network itself can
reveal a transformation from the intuitive choice of space to an optimal one. Evaluating in this space
should further improve performance of these models. We leave this for future work.

Ultimately, a good representation is one that makes a subsequent learning task
easier.
For unsupervised similarity tasks, this essentially reduces to how well the
model separates objects in the chosen representation space, and how appropriately the
similarity measure compares objects in that space.
Our findings lead us to the following practical advice: i) Use a simple model architecture where the optimal representation
space is clear by construction, or ii) use an arbitrarily complex model architecture and analyse the objective function to reveal,
for a chosen vector representation, an appropriate similarity metric.

We hope that future work will utilise a careful understanding of what similarity
means and how it is linked to the objective function, and that our analysis
can be applied to help boost the performance of other complex models.

\bibliography{library}
\bibliographystyle{iclr2018_workshop}

\newpage

\appendix

\section{Comparison with SkipThought}
\label{sec:appendix-decoder}

\begin{table}[t]
\centering
\small  
\caption{\small
  Performance of the SkipThought model, with and without layer normalisation \citep{Kiros2015, Ba2016},
  compared against the \gls{rnn}-\gls{rnn} model used in our experimental setup.
  On each task, the highest performing model is highlighted in bold.
  For SICK-R, we report the Pearson correlation, and for STS14 we
  report the Pearson/Spearman correlation with human-provided scores. For all
  other tasks, reported values indicate test accuracy.
  $\dagger$ indicates results taken from \cite{Conneau2017}.
  $\ddagger$ indicates our results from running SentEval on the model downloaded
  from \cite{Ba2016}'s publicly available codebase (\texttt{https://github.com/ryankiros/layer-norm}).
  We attribute the discrepancies in performance to differences in experimental setup or implementation.
  However, we expect our unrolling procedure to also boost SkipThought's performance on unsupervised
  similarity tasks, as we show for \gls{rnn}-\gls{rnn} in our fair single-codebase comparisons in the main text.
}
\label{tab:skipthought_original_comparison}
\resizebox{\textwidth}{!}{
  \begin{tabular}{c|cccccccccc}
\hline
\bf Model      & \bf MR   & \bf CR   & \bf SUBJ & \bf MPQA & \bf SST  & \bf TREC & \bf MRPC & \bf SICK-R & \bf SICK-E & \bf STS14 \\
\hline \hline
SkipThought $\dagger$    &     76.5 &     80.1 & 93.6     & 87.1     & 82.0     & \bf 92.2 & 73.0 & \bf 0.86   & \bf 82.3       & 0.29/0.35 \\
SkipThought-LN $\dagger$ & \bf 79.4 & \bf 83.1 & \bf 93.7 & \bf 89.3 & 82.9 & 88.4     & -        & \bf 0.86   & 79.5       & \bf 0.44/0.45 \\
SkipThought-LN $\ddagger$ & 78.6 & 82.2 & 92.9 & 89.1 & \bf 83.8 & 87.0  & \bf 73.2 & \bf 0.86 & 81.2 & 0.41/0.40\\
RNN-RNN        &     77.1 &     81.8 & 92.6     & 88.6    & 82.7     & 86.6     & 71.9     & 0.83       & 81.1       & 0.35/0.35 \\
\hline 
\end{tabular}}
\vspace{-0.75em}
\end{table}

See \autoref{tab:skipthought_original_comparison} for a comparison of our
\gls{rnn}-\gls{rnn} results with results of SkipThought from the literature.

\section{Cosine similarity on \gls{sts} tasks}
\label{sec:appendix-cos}

 \begin{table}[t]
\centering
\small  
\caption{\small
  Performance of different architectures and sentence representations on
  unsupervised similarity tasks using the cosine similarity between two vectors
  as the measure of their similarity.
  On each task, the highest performing setup for each
  encoder type is highlighted in bold and the highest performing setup overall
  is underlined.
  All reported values indicate Pearson/Spearman correlation
  coefficients for the task.
  \textbf{RNN encoder:} Using the raw encoder output (\gls{rnn}-\gls{rnn}) achieves the
  lowest performance across all tasks.
  Unrolling the \gls{rnn} decoders dramatically improves the performance
  across all tasks compared to using the raw encoder \gls{rnn} output, validating the
  theoretical justification presented in
  \autoref{subsec:similarity-sequence_decoder}. 
  \textbf{\gls{bow} encoder:} We do not observe the same uplift in performance from
  unrolling the \gls{rnn} encoder compared to the encoder output; fully understanding
  this behaviour is left for future work.
  }
\label{tab:similarity}
\resizebox{\textwidth}{!}{
\begin{tabular}{cc|ccccc}
\hline
 \textbf{Encoder} & \textbf{Decoder}   &                     \textbf{STS12}                      &                     \textbf{STS13}                      &                     \textbf{STS14}                      &                     \textbf{STS15}                      &                     \textbf{STS16}                      \\
  \hline\hline
&  BOW &             $\mathbf{0.466}/\mathbf{0.496}$             &                      $0.376/\mathbf{0.414}$                      &                 $\mathbf{0.478}/\mathbf{0.482}$                  &                      $0.424/0.454$                      &             $\mathbf{0.552}/\mathbf{0.586}$             \\
RNN & RNN &                      $0.323/0.357$                      &                      $0.320/0.319$                      &                      $0.345/0.345$                      &                      $0.402/0.409$                      &                      $0.373/0.408$                      \\
 & RNN-concat     &             $0.419/0.445$             &             $\mathbf{0.426}/\mathbf{0.414}$             &             $0.466/0.452$             &             $\mathbf{0.497}/\mathbf{0.503}$             &             $0.511/0.529$             \\
  \hline
  & BOW &                      $0.497/0.517$                      & $\underline{\mathbf{0.526}}/\underline{\mathbf{0.520}}$ &                      $\underline{\mathbf{0.576}}/0.561$                      &                      $0.604/0.605$                      &                      $\underline{\mathbf{0.592}}/\underline{\mathbf{0.592}}$                      \\

  BOW & RNN    &                      $0.508/0.526$                      &                      $0.483/0.489$                      &                      $0.575/\underline{\mathbf{0.562}}$                      & $\underline{\mathbf{0.644}}/\underline{\mathbf{0.641}}$ &                      $0.585/0.585$                      \\
     & RNN-concat     &             $\underline{\mathbf{0.521}}/\underline{\mathbf{0.540}}$             &             $0.491/0.498$             &             $0.561/0.554$             &             $0.627/0.625$             &             $0.584/0.581$             \\
  \hline 
\end{tabular}}
\vspace{-0.75em}
\end{table}

\begin{figure}[t]
  \centering
    
    {\includegraphics[width=\textwidth]{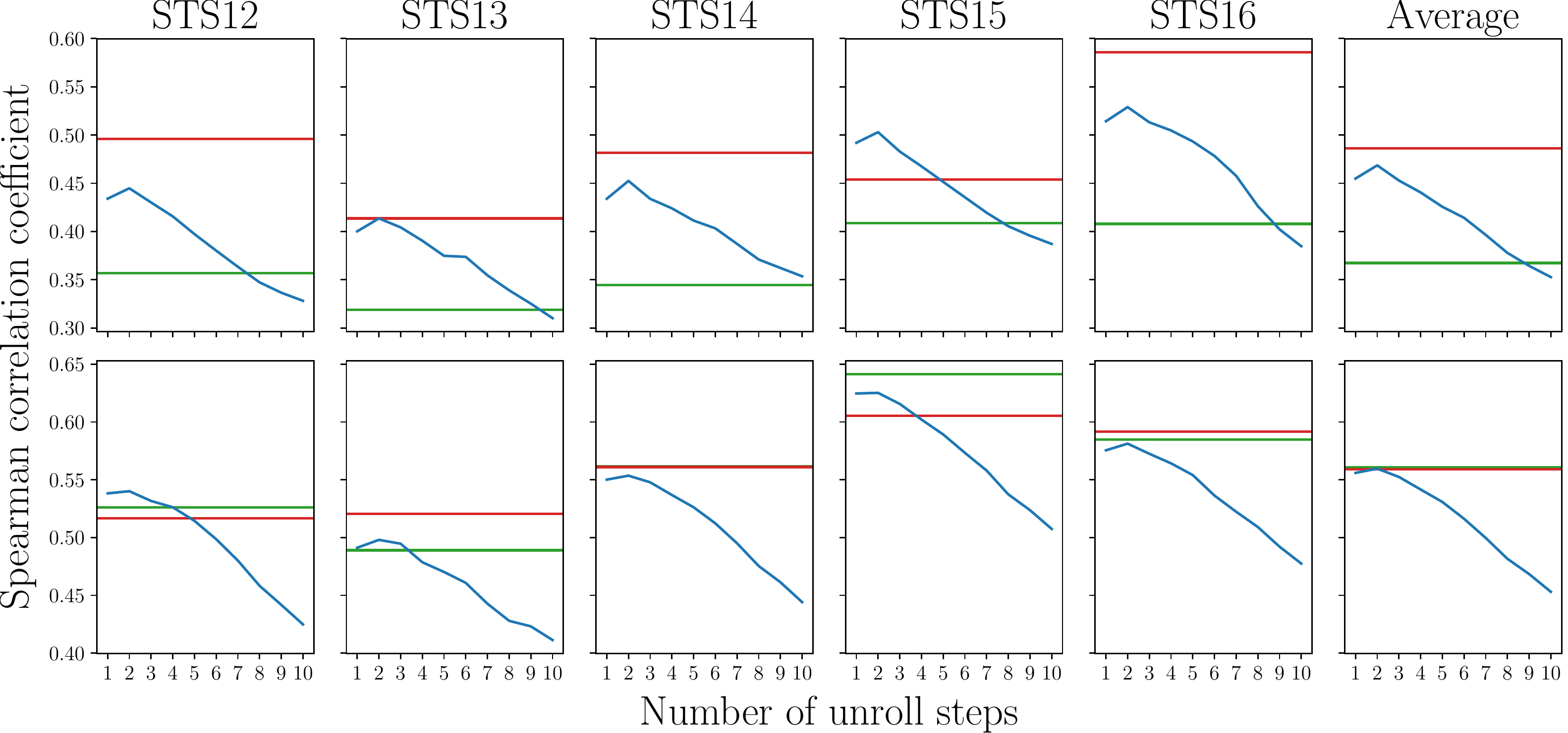}}
    \caption{\small Performance on the STS tasks depending on the number of unrolled hidden states of the decoders,
    using cosine similarity as the similarity measure.
    The top row presents results for the \gls{rnn} encoder and the bottom row for the \gls{bow} encoder.
    \textbf{Red:} Raw encoder output with \gls{bow} decoder. \textbf{Green:} Raw encoder output with \gls{rnn} decoder.
    \textbf{Blue:} Unrolled \gls{rnn} decoder output.
    For both \gls{rnn} and \gls{bow} encoders, unrolling the decoder
    strictly outperforms *-\gls{rnn} for almost every number of unroll steps, and perform nearly as well as or better than *-\gls{bow}.
    }
    \label{fig:decoder-unroll-cos}
    \vspace{-1em}
\end{figure}

As discussed in \autoref{sec:similarity}, the objective function
is maximising the dot product between the context and our stated optimal representations
(encoder output in the case of the \gls{bow} decoder,
and unrolled \gls{rnn} representation in the case of the \gls{rnn} decoder).
However, as other researchers in the field frequently
use cosine similarity for the \gls{sts} tasks, we present the results using cosine similarity in \autoref{tab:similarity}
and the results for different numbers of unrolled hidden decoder states in \autoref{fig:decoder-unroll-cos}.

Although the results in \autoref{tab:similarity} are mostly consistent with the dot product results in
\autoref{tab:similarity-dot}, the overall performance across \gls{sts} tasks is
noticeably lower when dot product is used instead of cosine similarity to
determine semantic similarity. Switching from using cosine similarity to dot
product transitions from considering only angle between two vectors, to also considering
their length. Empirical studies have indicated that the length of a word vector
corresponds to how sure of its context the model that produces it is. This
is related to how often the model has seen the word, and how many different
contexts it appears in (for example, the word vectors for ``January'' and
``February'' have similar norms, however, the word vector for ``May'' is
noticeably smaller) \citep{Schakel2015}.

A corollary is that longer sentences on
average have shorter norms, since they contain more words which, in turn, have
appeared in more contexts \citep{Adi2017}. During training, the corpus can
induce differences in norms in a way that strongly penalises sentences
potentially containing multiple contexts, and consequently will disfavour these
sentences as similar to other sentences under the dot product. This
potentially renders the dot product a less useful metric to choose
for \gls{sts} tasks than cosine similarity, which is unaffected by this issue.

\section{Unrolling RNN decoders by taking the mean}
\label{sec:appendix-mean}

A practical downside of the unrolling procedure described in \autoref{subsec:similarity-sequence_decoder} is that
concatenating hidden states of the decoder leads to very high dimensional vectors, which might be undesirable
due to memory or other practical constraints.  An alternative is to instead average the hidden states, which also
corresponds to a representation space in which the training objective optimises the dot product as a measure of
similarity between a sentence and its context.  We refer to this model choice as *-RNN-mean.

Results on similarity and transfer tasks for \gls{bow}-\gls{rnn}-mean and \gls{rnn}-\gls{rnn}-mean are presented
in \autoref{tab:similarity-dot-mean} and \ref{tab:transfer-mean} respectively, with results for the other models
from \autoref{subsec:experiments-results} included for completeness.  While the strong performance of \gls{rnn}-\gls{rnn}-mean
relative to \gls{rnn}-\gls{rnn} is consistent with our theory, exploring why it is able to outperform \gls{rnn}-concat experimentally
on \gls{sts} tasks is left to future work.


\begin{table}[t]
\centering
\small  
\caption{\small Performance of different architectures and sentence representations on
  unsupervised similarity tasks using dot product as the similarity measure.
  On each task, the highest performing setup for each
  encoder type is highlighted in bold and the highest performing setup overall
  is underlined.
  All reported values indicate Pearson/Spearman correlation
  coefficients for the task.
}
\label{tab:similarity-dot-mean}
\resizebox{\textwidth}{!}{
\begin{tabular}{cc|ccccc}
\hline
  \textbf{Encoder} & \textbf{Decoder}  &                     \textbf{STS12}                      &                     \textbf{STS13}                      &                     \textbf{STS14}                      &                     \textbf{STS15}                      &           \textbf{STS16}           \\
\hline \hline
     & BOW      &                 $0.286/\mathbf{0.384}$                  &                      $0.381/0.426$                      &                      $0.365/0.406$                      &                      $0.262/0.392$                      &           $0.260/0.328$            \\
     RNN & RNN      &                      $0.267/0.286$                      &                      $0.371/0.365$                      &                      $0.357/0.348$                      &                      $0.379/0.379$                      &           $0.313/0.303$            \\
   & RNN-mean   &                      $0.330/0.361$                      &             $\mathbf{0.420}/\mathbf{0.427}$             &             $\mathbf{0.438}/\mathbf{0.428}$             &             $\mathbf{0.419}/\mathbf{0.426}$             &           $0.324/0.342$            \\
  & RNN-concat  &                 $\mathbf{0.335}/0.359$                  &                      $0.411/0.415$                      &                      $0.413/0.403$                      &                      $0.414/0.419$                      &  $\mathbf{0.326}/\mathbf{0.347}$   \\ \hline
     & BOW      &                      $0.351/0.390$                      &                      $0.418/0.478$                      &                      $0.442/0.488$                      &                      $0.455/0.535$                      & $0.370/\underline{\mathbf{0.461}}$ \\
     BOW & RNN      &                      $0.310/0.342$                      &                      $0.365/0.422$                      &                      $0.396/0.440$                      &                      $0.412/0.469$                      &           $0.281/0.330$            \\
   & RNN-mean   &                      $0.394/0.414$                      &                      $0.469/0.495$                      &                      $0.490/0.498$                      &                      $0.495/0.530$                      &           $0.381/0.439$            \\
  & RNN-concat  & $\underline{\mathbf{0.422}}/\underline{\mathbf{0.438}}$ & $\underline{\mathbf{0.478}}/\underline{\mathbf{0.498}}$ & $\underline{\mathbf{0.498}}/\underline{\mathbf{0.506}}$ & $\underline{\mathbf{0.512}}/\underline{\mathbf{0.544}}$ & $\underline{\mathbf{0.402}}/0.460$ \\
\hline
\end{tabular}}
\vspace{-0.75em}
\end{table}


\begin{table}[t]
\centering
\caption{\small Performance of different architectures and sentence representations on
  supervised transfer tasks. On each task, the highest performing setup for each
  encoder type is highlighted in bold and the highest performing setup overall
  is underlined. All reported values indicate test accuracy on the task, except
  for SICK-R where we report the Pearson correlation with human-provided
  scores.
}
\label{tab:transfer-mean}
\small
\resizebox{\textwidth}{!}{
\begin{tabular}{cc|ccccccccc}
\hline
  \textbf{Encoder} & \textbf{Decoder}   &         \textbf{MR}          &         \textbf{CR}          &        \textbf{MPQA}         &        \textbf{SUBJ}         &         \textbf{SST}         &        \textbf{TREC}         &        \textbf{MRPC}         &       \textbf{SICK-R}       &       \textbf{SICK-E}        \\
\hline\hline
    & BOW      &           $75.78$            &           $79.34$            &           $86.25$            &           $90.77$            &           $81.99$            &           $84.60$            &           $70.55$            &           $0.80$            &           $78.81$            \\
   RNN & RNN    & $\mathbf{77.06}$ &       $81.77$       & $\mathbf{88.59}$ &       $\mathbf{92.56}$       & $\underline{\mathbf{82.65}}$ &       $86.60$       &           $71.94$            &           $0.83$            & $\mathbf{81.10}$ \\
                   & RNN-mean &           $76.55$            &           $81.03$            &           $87.35$            &           $92.29$            &           $81.11$            &           $84.80$            & $\mathbf{73.51}$ &       $\mathbf{0.84}$       &           $78.22$            \\
  & RNN-concat &           $76.20$            &           $\textbf{82.07}$            &           $85.96$            &           $91.80$            &           $80.83$            &           $\textbf{87.20}$            & $71.59$ &       $0.82$       &           $80.35$            \\ \hline
 &    BOW      &       $76.16$       &           $81.14$            &       $87.03$       & $92.77$ &       $81.66$       &           $84.20$            &           $71.07$            & $0.84$ &       $80.58$       \\
  BOW &  RNN    &           $76.05$            & $\underline{\mathbf{82.07}}$ &           $85.80$            &           $92.13$            &           $80.83$            & $87.20$ &       $72.99$       &           $0.82$            &           $78.87$            \\
 & RNN-mean &           $75.85$            &           $81.30$            &           $85.54$            &           $90.80$            &           $80.12$            &           $84.00$            &           $71.13$            &           $0.81$            &           $77.76$            \\
  & RNN-concat &           $\underline{\textbf{77.27}}$            &           $82.04$            &           $\underline{\textbf{88.74}}$            &           $\underline{\textbf{92.88}}$            &           $\textbf{81.82}$            &           $\underline{\textbf{89.60}}$            & $\underline{\mathbf{73.68}}$ &       $\underline{\mathbf{0.85}}$       &           $\underline{\mathbf{82.26}}$            \\ \hline
\end{tabular}}
\vspace{-0.75em}
\end{table}



%

\end{document}